\documentclass[conference]{IEEEtran}

\usepackage{bm}

\usepackage{todonotes}
\usepackage{caption}

\newcommand\blfootnote[1]{%
  \begingroup
  \renewcommand\thefootnote{}\footnote{#1}%
  \addtocounter{footnote}{-1}%
  \endgroup
}

\usepackage{xargs}	                
\newcommandx{\mattia}[2][1=]{\todo[backgroundcolor=green!25,inline,#1]{MR: #2}\noindent}
\newcommandx{\hakan}[2][1=]{\todo[backgroundcolor=blue!25,inline,#1]{HG: #2}\noindent}
\newcommandx{\yongjoon}[2][1=]{\todo[backgroundcolor=orange!45,inline,#1]{YJ: #2}\noindent}
\newcommandx{\jeremy}[2][1=]{\todo[backgroundcolor=red!45,inline,#1]{JM: #2}\noindent}

\newcommand{\eg}[0]{\textit{e.g.}, }
\newcommand{\ie}[0]{\textit{i.e.}, }

\definecolor{Siena}{RGB}{180,72,25}
\newcommandx{\odd}[2][1=]{\textcolor{Siena}{#2}} 

\usepackage{subfig}
\usepackage[normalem]{ulem}
\usepackage{booktabs}
\usepackage{url}
\usepackage{xcolor}
\usepackage{tabu}
\usepackage{colortbl}

\usepackage{adjustbox}
\usepackage{tablefootnote}

\begin{document}

\title{Online and Offline Robot Programming via Augmented Reality Workspaces}

\author{\IEEEauthorblockN{Yong Joon Thoo,
Jérémy Maceiras,
Philip Abbet,
Mattia Racca,
Hakan Girgin,
and Sylvain Calinon}}

\maketitle

\begin{abstract}\blfootnote{The authors are with the Idiap Research Institute, Martigny, Switzerland (name.surname@idiap.ch).\vskip.2em
This work was supported by the COLLABORATE project (https://collaborate-project.eu), funded by the EU within H2020-DT-FOF-02-2018 under grant agreement 820767, by the LEARN-REAL project (https://learn-real.eu, CHIST-ERA), funded by the Swiss National Science Foundation, and by the ROSALIS project, funded by the Swiss National Science Foundation.}
Robot programming methods for industrial robots are time consuming and often require operators to have knowledge in robotics and programming. To reduce costs associated with reprogramming, various interfaces using augmented reality have recently been proposed to provide users with more intuitive means of controlling robots in real-time and programming them without having to code. However, most solutions require the operator to be close to the real robot's workspace which implies either removing it from the production line or shutting down the whole production line due to safety hazards. We propose a novel augmented reality interface providing the users with the ability to model a virtual representation of a workspace which can be saved and reused to program new tasks or adapt old ones without having to be co-located with the real robot. Similar to previous interfaces, the operators then have the ability to program robot tasks or control the robot in real-time by manipulating a virtual robot. We evaluate the intuitiveness and usability of the proposed interface with a user study where 18 participants programmed a robot manipulator for a disassembly task.
\end{abstract}

\IEEEpeerreviewmaketitle

\begin{figure}[!t]
\centering
    \subfloat[\label{fig_frontWorkspace}]{\includegraphics[width=1.5in]{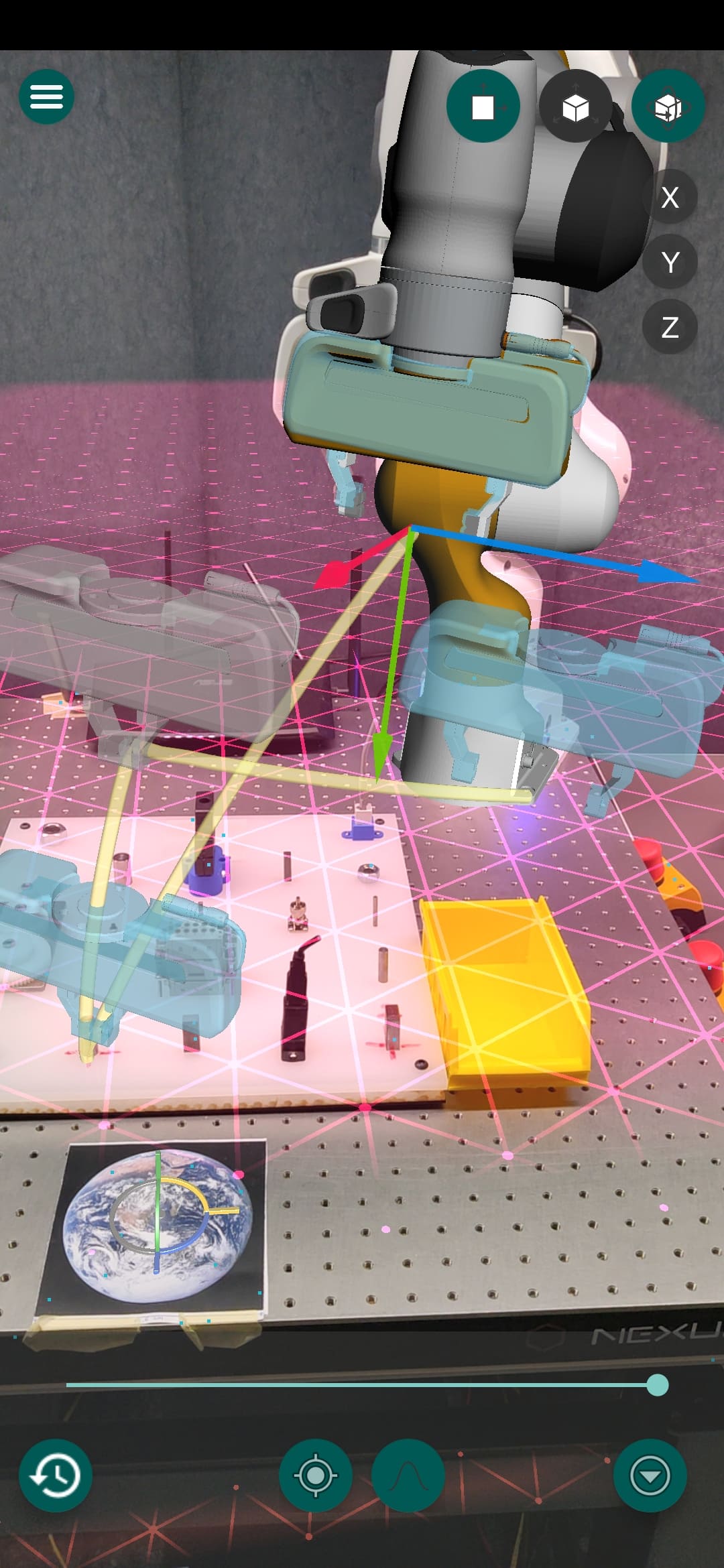}}
    \hfill
    \subfloat[\label{fig_frontVirtual}]{\includegraphics[width=1.5in]{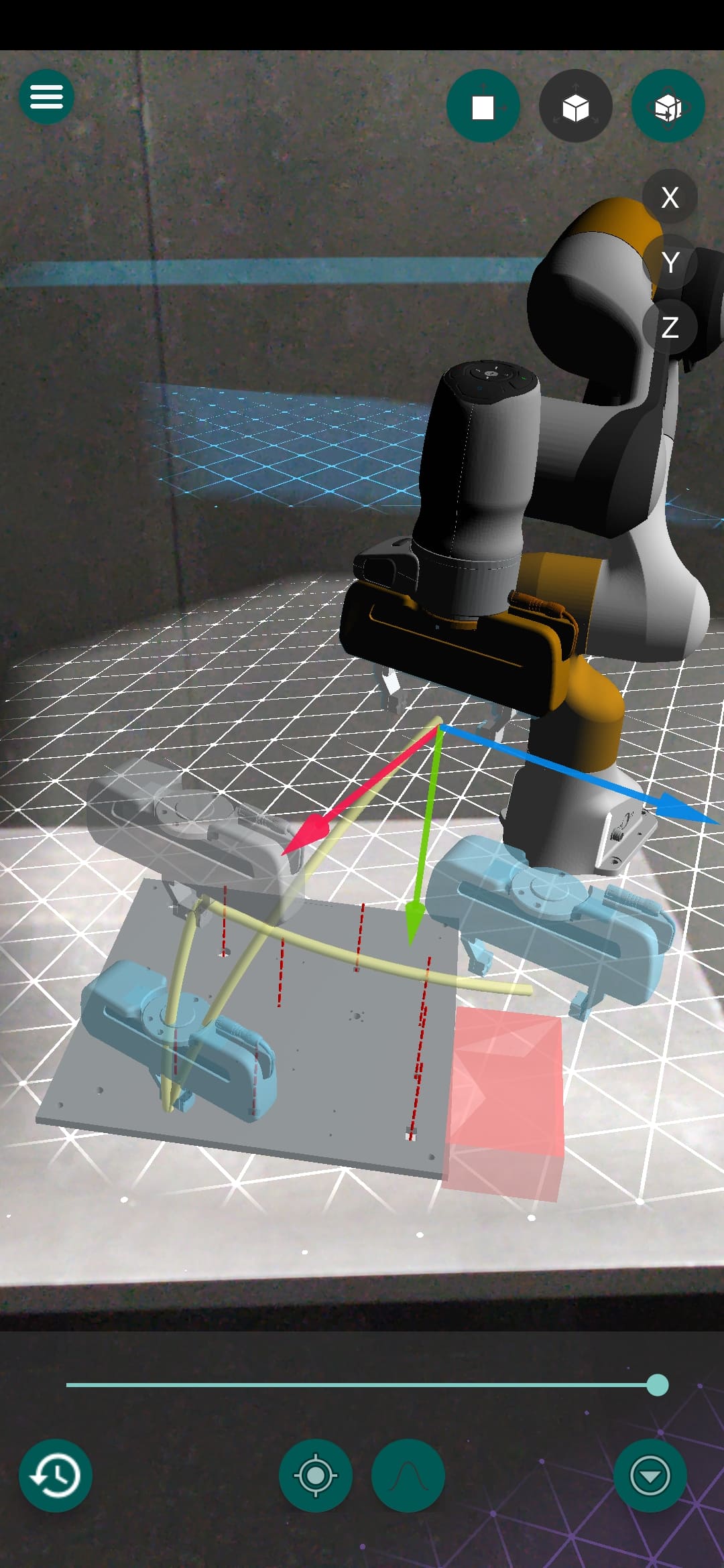}}
    \caption{Illustration of the ability to program the robot \emph{offline} either (a) directly in the real robot's workspace or (b) in a virtual workspace by defining keypoints (displayed as transparent end-effectors) and visualising the resulting trajectory.}
    \label{fig_frontPage}
\end{figure}

\section{Introduction}

Industrial robots play a vital role in manufacturing tasks such as welding and assembly due to their ability to perform these tasks with a high degree of precision and reliability. Nevertheless, despite their importance in manufacturing, the programming of such robots remains a costly and time-consuming task often requiring operators to have a certain degree of knowledge of robotics and programming \cite{IEEEfull:Calinon_2019, IEEEfull:Zhang_2020, IEEEfull:Blankemeyer_2018, IEEEfull:Gradmann_2018, IEEEfull:Chacko_2020, IEEEfull:Abbas_2012}.

The most popular methods for programming industrial robots are referred to as \emph{online} and \emph{offline}. The \emph{online} method requires the use of the real physical robot and consists of recording the path designed by an operator by controlling the joints of the robot via a teaching pendant with a joystick, a 6D-mouse, or a keypad. The \emph{offline} method on the other hand does not require the actual robot and consists of constructing a virtual representation of the robot and its workspace to simulate tasks prior to applying it on the real robots \cite{IEEEfull:Ong_2020, IEEEfull:Ostanin_2020}. Various simulators and 3D engines have recently been extended to support robotics platforms, making the \emph{offline} programming of robots possible.

These methods present however multiple disadvantages.
For \emph{online} programming, the control of a robot via a teaching pendant often requires the robot in question to be removed from production during programming or, under certain circumstances, a shutdown of the production line \cite{IEEEfull:Zhang_2020}.
In the case of \emph{offline} programming, the reconstruction of the robot's workspace in a virtual environment can be time consuming.
Furthermore, the transition from the simulation to the execution on the real robot may suffer from inaccuracies, with additional costs required to overcome them \cite{IEEEfull:Ong_2020}.
Moreover, offline methods require a certain degree of familiarity with programming and simulators.

The significance of these disadvantages increases when robots need to be reprogrammed frequently. This is especially true in High-Mix Low-Volume (HMLV) manufacturing, where small quantities of a large variety of items are produced~\cite{IEEEfull:Ong_2020}.
As such, research has been conducted to design intuitive and efficient programming interfaces for users with little to no experience in the fields of robotics or programming \cite{IEEEfull:Calinon_2019}.

To suit these needs, Augmented Reality (AR) technology has been proposed as a solution due to its ability to superimpose information in various forms onto the real world.
Indeed, AR has been growing increasingly popular in recent years and has been used in various fields such as architecture, construction, education, manufacturing and engineering \cite{IEEEfull:Chi_2013}.
In the field of robotics, AR interfaces provide a new medium for interaction with the robot and enable the exchange of information during tasks \cite{IEEEfull:Makhataeva_2020} in addition to providing users with the ability to preview a robot's intended actions.

Several AR or Mixed Reality (MR) interfaces have been proposed to enable users not only to control a robot \cite{IEEEfull:Abbas_2012} but also to program it by defining trajectories and tasks via the manipulation of a virtual robot in the interface \cite{IEEEfull:Chacko_2020, IEEEfull:Ostanin_2020, IEEEfull:Quintero_2018, IEEEfull:Ostanin_2018, IEEEfull:Gadre_2019}.
These interfaces provide the operators with an intuitive way of programming the robot without an extensive knowledge about programming and combines some of the advantages of current \emph{online} and \emph{offline} methods by enabling users to simulate a task directly in the robot's workspace, making the transition from simulation to real execution easier. However, the proposed solutions require the users to share the same workspace as the robot such that the robot in question cannot be used for production during the programming process.

To further reduce the time during which the robots are unavailable for production, we propose an AR interface providing users with the ability to program the robot from a virtual representation of the workspace which can be built and saved by the user such that tasks can be programmed from different locations (Fig. \ref{fig_frontVirtual}). Adjustments to the trajectory, if necessary, could then be made by displaying the designed task in the real robot's workspace (Fig. \ref{fig_frontWorkspace}) resulting in less time during which the robot in question would be unavailable for production. 

In Section~\ref{section:related_work}, we first present a state of the art of robot control and programming via augmented reality. This is followed by a description of our proposed AR interface and its components in Section \ref{section:methodology}. In Section~\ref{section:experiments}, we describe the experiments conducted to evaluate our interface. In Section~\ref{section:results}, we present and give a brief discussion on the results of our experiments. Finally, we conclude in Section~\ref{section:conclusion} with some potential future work descriptions.

\section{Related work}
\label{section:related_work}

\begin{figure}[!t]
\centering
    \subfloat[\label{fig_manualPlacing}]{\includegraphics[width=1.5in]{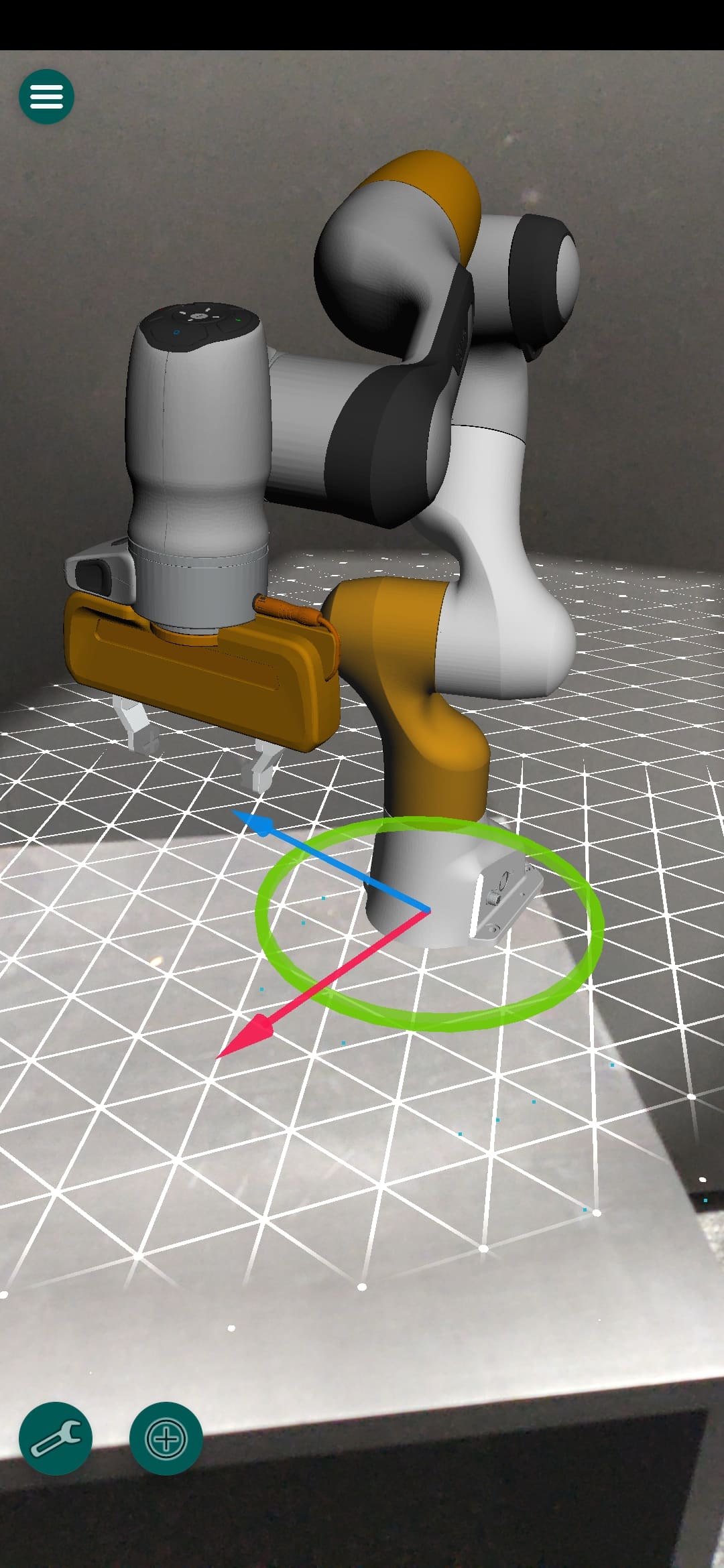}}
    \hfill
    \subfloat[\label{fig_markerCalib}]{\includegraphics[width=1.5in]{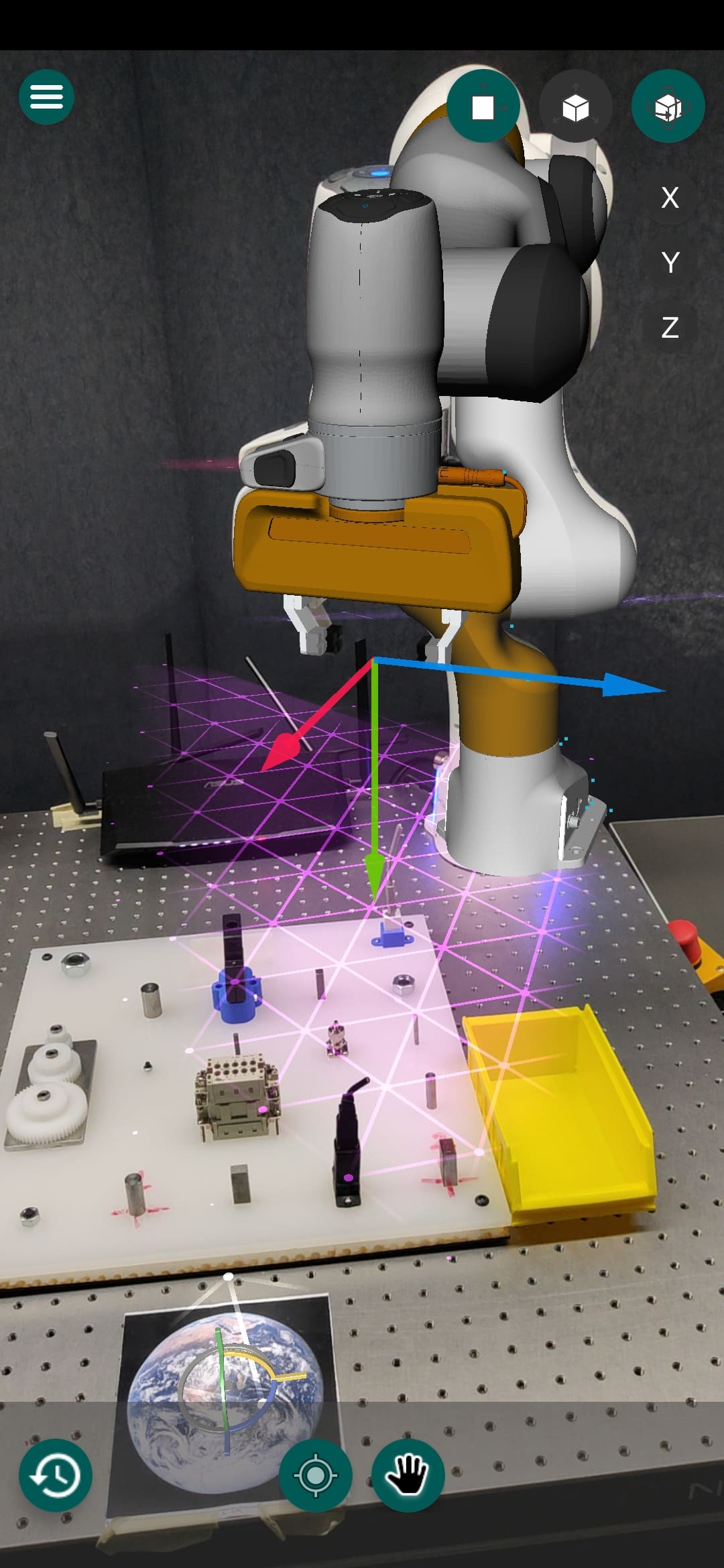}}
    \caption{Illustration of the methods to place a robot in the scene: (a) \emph{manual} and (b) \emph{marker calibration} (here, with a picture of Earth).}
    \label{fig_robotPlacing}
\end{figure}

AR has been used in combination with robotics for various use cases allowing users to not only interact with or control the robot and visualise their intentions but also program the robot by defining target points in the actual workspace that the robot should reach. This enables inexperienced users with little to no programming skills to design complex robot operations and ensure that the trajectory to be taken by the real robot is collision free \cite{IEEEfull:Zhang_2020}. Various AR interfaces displayed either via a Head Mounted Device (HMD), smartphone, tablet or screen have been proposed in recent years for industrial settings as well as for collaborative tasks where the user can perform the task alongside the operator.

In \cite{IEEEfull:Ostanin_2020, IEEEfull:Quintero_2018, IEEEfull:Ostanin_2018, IEEEfull:Gadre_2019}, the authors present mixed reality interfaces with a Microsoft HoloLens to program tasks by defining waypoints directly in the robot's workspace and previewing the trajectory via a virtual robot before sending the commands to a real robot. To place the waypoints and interact with the other features each respective interface offers, the interfaces employ features available on the HoloLens, namely gesture recognition as well as speech recognition and head pose in some cases \cite{IEEEfull:Quintero_2018}. The main drawbacks of such interfaces are that the selection and dragging features in the HoloLens may not always be reliable due to imperfect hand tracking \cite{IEEEfull:Gadre_2019} such that the trajectory may not be optimal and it may be difficult to perform tasks that demand a high degree of precision. To overcome this issue, in \cite{IEEEfull:Ostanin_2020}, the authors present an interface providing users with the ability to scale the path for a more accurate planning during such tasks, as well as the selection of the method by which the trajectory between waypoints should be interpolated (line, arc, etc.). However, despite allowing users to be hands-free, the HoloLens may be less intuitive than a 2D interface for novice users and HMDs are not necessarily available to most users. Additionally, it has been reported that such devices may cause discomfort and sickness which may be affecting their industrial acceptance \cite{IEEEfull:Chacko_2019}.

In recent years, various platforms and toolkits such as ARCore, ARToolKit and Mixed Reality Toolkit (MRTK) have been introduced allowing users to develop augmented or mixed reality applications for certain smartphones and tablets making the technology more accessible. Such toolkits have been used to develop interfaces enabling users to perform pick-and-place tasks in collaboration with a robot \cite{IEEEfull:Chacko_2019}, define waypoints on a 2D surface for a tool to pass through \cite{IEEEfull:Chacko_2020} as well as control the joints of the robot via the interface, and visualise a trajectory demonstrated to the robot via kinesthetic teaching \cite{IEEEfull:Gradmann_2018}.
However, in the case of \cite{IEEEfull:Chacko_2019, IEEEfull:Chacko_2020}, the interfaces do not provide a visualisation of the whole robot which may be important if the operator needs to check for eventual collisions. Whilst this is not the case with the interface presented in \cite{IEEEfull:Gradmann_2018}, the teaching mode presented in the interface involves demonstrating a task to the robot such that the robot is backdrivable through a gravity compensation controller, which is not available to most industrial robots.

In addition to some of the drawbacks mentioned above, most of the interfaces presented in this section require the operators to employ the interface whilst being next to the robot's workspace where the operators can interact directly with the physical objects present in the environment. Due to safety issues, similar to teaching a robot a task via a pendant, this would require shutting down the robots in the production line. 
Table~\ref{comparisonTable} provides a summary of traditional and state of the art interfaces for programming industrial robots, along with their advantages and drawbacks.

In this paper, we propose an AR interface available on smartphone and tablet providing users with the ability to not only control and program a robot in its workspace but also model the workspace via virtual objects. This provides the advantage of being able to use the interface to program robot tasks from different locations and adapt previously saved tasks to new situations,  environments, tools, or robots.

\begin{table*}[!t]
\centering
\caption{Comparison table of traditional and state of the art interfaces for industrial robot programming.}
\label{comparisonTable}
\adjustbox{max width=\textwidth}{\begin{tabular}{|l|l|m{5.75cm}|m{5.75cm}|}
\hline
\rowcolor[HTML]{D0D0D0} & & & \\[.5ex] \rowcolor[HTML]{D0D0D0} 
\textbf{} & \textbf{Description} & \textbf{Advantages} & \textbf{Disadvantages} \\ \hline
\textbf{\begin{tabular}[c]{@{}l@{}}Teach pendant\\ interfaces\end{tabular}} & \begin{tabular}[c]{@{}l@{}}Direct control via a joystick or 6D\\ mouse for point-to-point or block-based\\ programming on the fly, \eg \\Kuka SmartPAD, Franka Emika Desk.\end{tabular} & \begin{itemize}
    \item[$+$] No need to model the environment, as the robot directly operates on it
    \item[$+$] Familiar interface by being the industry standard
\end{itemize} & \begin{itemize}
    \item[$-$] Little to no integration with external sensors
    \item[$-$] Requires co-presence of the user with the robot or additional visualisation interface if remote teleoperation is employed
    \item[$-$] No inspection for safety of the motion prior to execution
    \item[$-$] Brand specific tool 
    \end{itemize}\\ \hline
\rowcolor[HTML]{D0D0D0} 
\textbf{\begin{tabular}[c]{@{}l@{}}Visualisation tools \\ and physics \\ simulators\end{tabular}} & \begin{tabular}[c]{@{}l@{}}Visualisation tools, e.g. RViz and\\ simulators \eg PyBullet, CoppeliaSim,\\ MuJoCo, Gazebo.\end{tabular} & \begin{itemize}
    \item[$+$] Online \& offline programming options are available
    \item[$+$] Visualisation of robot model, robot programs, and additional sensor data
    \item[$+$] Not bound to a specific robot platform\end{itemize} & \begin{itemize}
        \item[$-$] Need to model the environment for offline programming
        \item[$-$] Transferring the planned motions requires accurate simulation of the robot, its sensors, and the environment (sim-to-real gap)
        \item[$-$] Selection of motion keypoints in the software requires accurate simulation of the environment (real-to-sim gap)\end{itemize} \\ \hline
\textbf{ABB RobotStudio\footnotemark} & \begin{tabular}[c]{@{}l@{}} Simulator that generates code for ABB \\ robots.\end{tabular} & \begin{itemize}
    \item[$+$] Monitoring of robot models, work cells and routines, either on desktop (via simulated environment) or with AR enabled devices 
\end{itemize} & \begin{itemize}
    \item[$-$] Depth perception issues linked to AR
    \item[$-$] Programming only via the desktop interface
    \item[$-$] Offline programming requires complete modelling of the environment
    \item[$-$] Requires accurate simulation of the environment (real-to-sim gap)
    \item[$-$] Brand specific tool
    \end{itemize}\\ \hline
\rowcolor[HTML]{D0D0D0} 
\textbf{\begin{tabular}[c]{@{}l@{}}Headset AR/MR\\ enabled solutions\end{tabular}} & \begin{tabular}[c]{@{}l@{}}Headset AR/MR enabled devices\\ \cite{IEEEfull:Ostanin_2020, IEEEfull:Quintero_2018, IEEEfull:Ostanin_2018, IEEEfull:Gadre_2019}.\end{tabular} & \begin{itemize}
    \item[$+$] Visualisation of robot model, robot programs and sensor data (ROS enabled for \cite{IEEEfull:Ostanin_2020, IEEEfull:Gadre_2019})
    \item[$+$] Multimodal input/output (speech, visual, tactile)
    \item[$+$] Offline programming and safety checks for online programming do not require environment modeling
    \item[$+$] Hands-free
    \item[$+$] Not bound to a specific robot platform
    \end{itemize} & \begin{itemize}
        \item[$-$] Depth perception issues linked to AR
        \item[$-$] Offline programming provides a visualisation of the virtual robot in the real scene but no other elements of the environment
        \item[$-$] Reported cases of discomfort and sickness\end{itemize} \\ \hline
\textbf{\begin{tabular}[c]{@{}l@{}}Mobile device \\ AR enabled \\ solutions\end{tabular}} & \begin{tabular}[c]{@{}l@{}}Mobile AR enabled devices (\eg tablet, \\ smartphone, $\dots$) \cite{IEEEfull:Gradmann_2018, IEEEfull:Chacko_2020, IEEEfull:Chacko_2019}.\end{tabular} & \begin{itemize}
    \item[$+$] Online \& offline programming options are available
    \item[$+$] Visualisation of robot model, robot programs, and additional sensor data
    \item[$+$] Multimodal input/output (speech, visual, tactile)
    \item[$+$] Offline programming and safety checks for online programming do not require environment modeling
    \item[$+$] Not bound to a specific robot platform\end{itemize} & \begin{itemize}
        \item[$-$] Depth perception issues linked to AR
        \item[$-$] Tactile screen can lead to inaccurate motions
        \item[$-$] Offline programming provides a visualisation of the virtual robot in the real scene but no other elements of the environment\end{itemize} \\ \hline
\rowcolor[HTML]{D0D0D0} 
\textbf{Proposed solution} & \begin{tabular}[c]{@{}l@{}}Interface displayed via a mobile AR\\ enabled device where users can\\ visualise a virtual robot and \\additional elements of the environment.\\ Tasks can be programmed\\ either directly in the real workspace or\\ in a virtual workspace.\end{tabular} & \begin{itemize}
    \item[$+$] Online \& offline programming options are available
    \item[$+$] Visualisation of robot model, robot programs, and sensor data (ROS enabled)
    \item[$+$] Offline programming and safety checks for online programming do not require environment modeling
    \item[$+$] Intuitive selection of the keypoints in the interface mitigates the real-to-sim gap problem
    \item[$+$] Not bound to a specific robot platform\end{itemize} & \begin{itemize}
        \item[$-$] Depth perception issues linked to AR
        \item[$-$] Tactile screen can lead to inaccurate motions\end{itemize} \\ \hline
\end{tabular}}
\end{table*}
\footnotetext{https://new.abb.com/products/robotics/robotstudio}

\section{Methodology}
\label{section:methodology}

\begin{figure}[!t]
\centering
\includegraphics[width=0.425\textwidth]{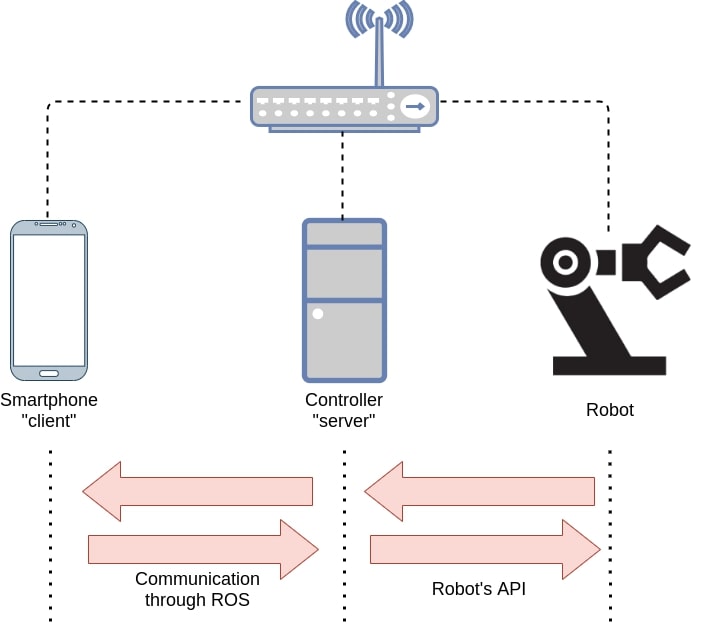}
\caption{Communication diagram of the application.}
\label{fig_communication}
\end{figure}

The interface proposed in this work was developed with Google's ARCore software library on Android Studio such that it can run on most AR compatible Android devices\footnote{https://developers.google.com/ar/devices}. The main functionality of the interface is that it provides non-expert users with an \emph{offline} programming solution by means of a virtual robot in a simulation environment, as well as an \emph{online} programming solution by controlling the real robot with the provided AR tools.

The virtual robot can be placed at a desired location within the environment and can then be connected to the real one, if required, to visualise what has been programmed offline or to directly control the robot online. We implemented several mechanisms to control the robot to achieve tasks in end-effector space. Additionally, the interface provides various options for the user to add/modify objects and obstacles, which can be moved, rotated and scaled in the workspace of the robot. The workspace created virtually can also be saved for later programming without requiring the real robot to be operated at the same time.

Videos of the AR interface and the experiments described in Section \ref{section:experiments} are available at: \url{https://sites.google.com/view/idiap-ar-robot-interface/}

\subsection{Simulation of the robot via ARCore}
We first describe the procedure to create a virtual robot in the desired workspace and how the connection to the real robot is established and calibrated.
Upon launch, ARCore detects planes which allow the users to interact with the virtual objects placed in the environment.
The virtual robot is spawned following the same procedure adopted by the Kinematics and Dynamics Library\footnote{https://www.orocos.org/kdl.html} (KDL).
Description in the Unified Robot Description Format (URDF) informs the creation of the kinematic chains representing the real robot.

While in this work we focus on the Franka Emika manipulator, our interface can readily handle any robot as long as a URDF file and related meshes are available.
The technological feasibility of adding robots to the interface should however be considered in the larger scope of usability, as characteristics of specific robotic platforms may not mesh well with the design of our interface.

\subsubsection{Positioning the virtual robot in the workspace}

The interface exposes two methods to place the virtual robot in the AR workspace: a \emph{manual} method and a \emph{marker calibration} method.
The \emph{manual} method lets the user select a location on a plane detected by ARCore upon which to place the virtual robot (as shown in Fig. \ref{fig_manualPlacing}).
Once linked to the detected plane, the robot's orientation and position can then be further adjusted with the blue and red translation axes and the green rotation ring, as shown in Fig.~\ref{fig_manualPlacing}.

The \emph{marker calibration} method exploits ARCore's Augmented Images APIs\footnote{https://developers.google.com/ar/develop/java/augmented-images} to superimpose the virtual robot onto the real robot as illustrated in Fig. \ref{fig_markerCalib}.
It first detects the pose of a previously placed calibration marker in the workspace with respect to the ARCore coordinate system.
Using a given homogeneous transformation between the calibration marker and the coordinate system at the base of the robot, the virtual robot is accordingly placed in the ARCore's interface.

\begin{figure*}[!t]
\centering
\subfloat[\label{fig_transAxis}]{\includegraphics[width=1.5in]{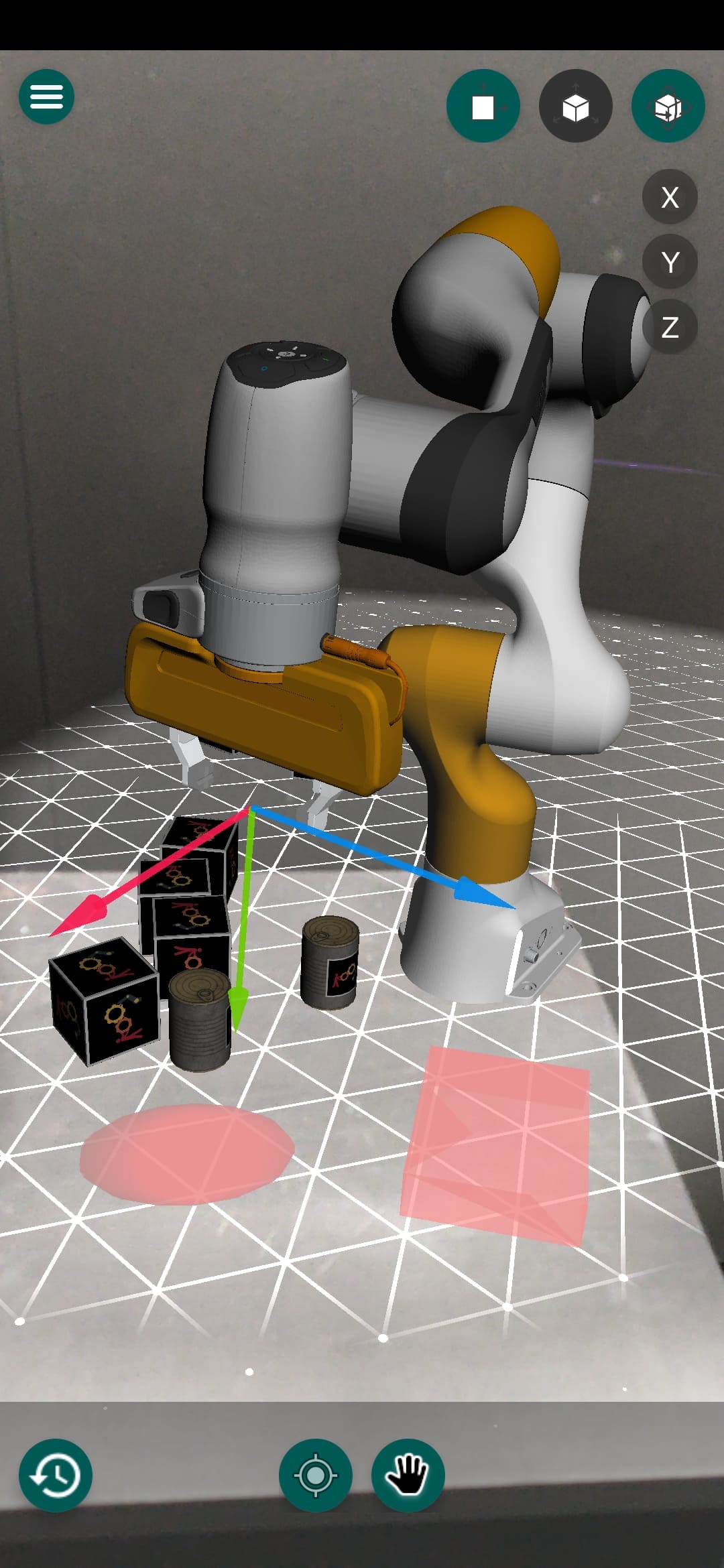}}
\hfill
\subfloat[\label{fig_rotRings}]{\includegraphics[width=1.5in]{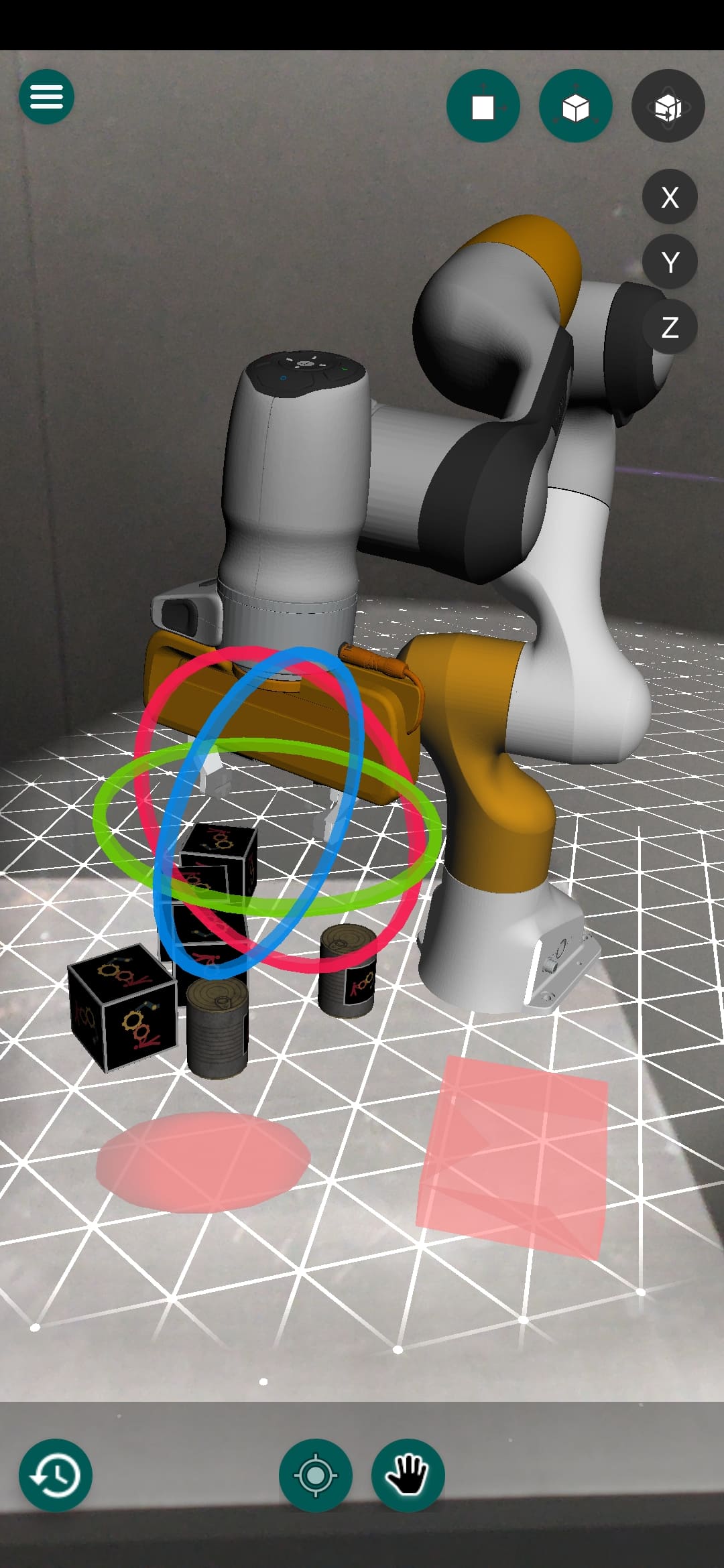}}
\hfill
\subfloat[\label{fig_planeMotion}]{\includegraphics[width=1.5in]{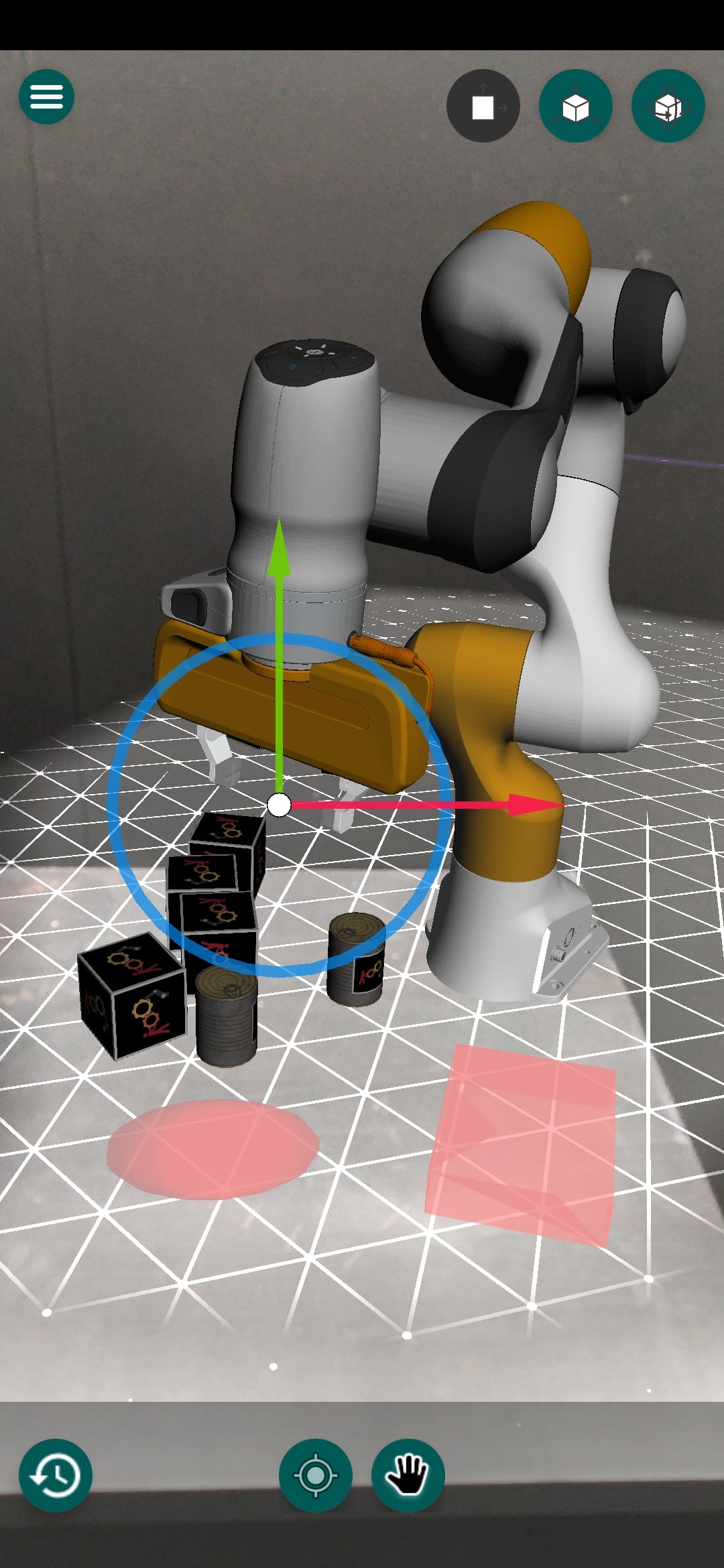}}
\hfill
\subfloat[\label{fig_elbowManip}]{\includegraphics[width=1.5in]{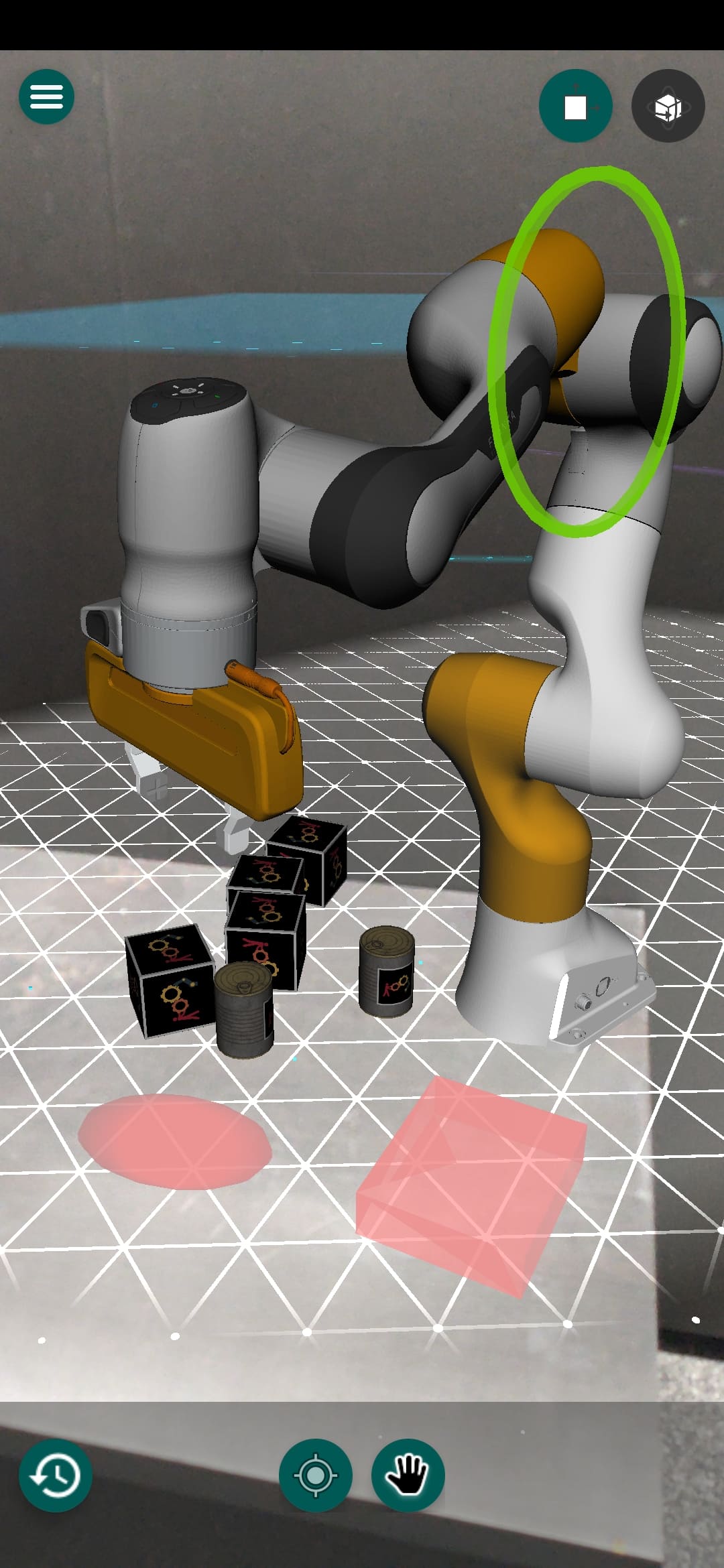}}
\caption{Fully virtual workspace with objects and obstacles (in red) and the various ways of controlling a robot: (a) control of the end-effector position, (b) control of the end-effector orientation, (c) control of the end-effector along a plane and (d) control of the elbow.}
\label{fig_virtComponents}
\end{figure*}

\subsubsection{Communication with the real robot}

The virtual robot is connected to the real robot via Robot Operating System\footnote{https://www.ros.org/} (ROS) which allows interactions between the real and virtual robots. As ROS is already available for most of the available industrial and collaborative robots, it provides a generic interface, enabling additionally the use of multiple sources of sensory information together for the robot to work with. Our interface thus allows the ARCore device to communicate with the server via ROS, as illustrated in Fig. \ref{fig_communication}-\emph{left}. The robot is controlled via ROS through a server enabling its real-time communication with the robot motors using the robot's API provided by the manufacturer of the robot, as illustrated in Fig. \ref{fig_communication}-\emph{right}. As the implementation of our interface is built on top of ROS and not on a specific robot's API, the proposed interface is robot agnostic.

\subsection{Motion control}
\label{section:controlTechniques}

We present here the methods available to control the robot's joints and end-effector as well as to plan trajectories. 

As opposed to joystick or teaching pendant based methodologies of programming robots, we provide more diverse options to control and program the robot.
Users are able to reposition the robot's end-effector using the translation axes shown in Fig. \ref{fig_transAxis} and orient it using the rotational rings, each describing the rotation around the translation axes, as shown in Fig. \ref{fig_rotRings}.
These changes in the task space are applied to the joint space through a weighted inverse kinematics algorithm.
Internally, a joint impedance controller was used to reach new joint space targets.

\begin{figure}[!t]
\centering
    \subfloat[\label{fig_exp2}]{\includegraphics[width=1.5in]{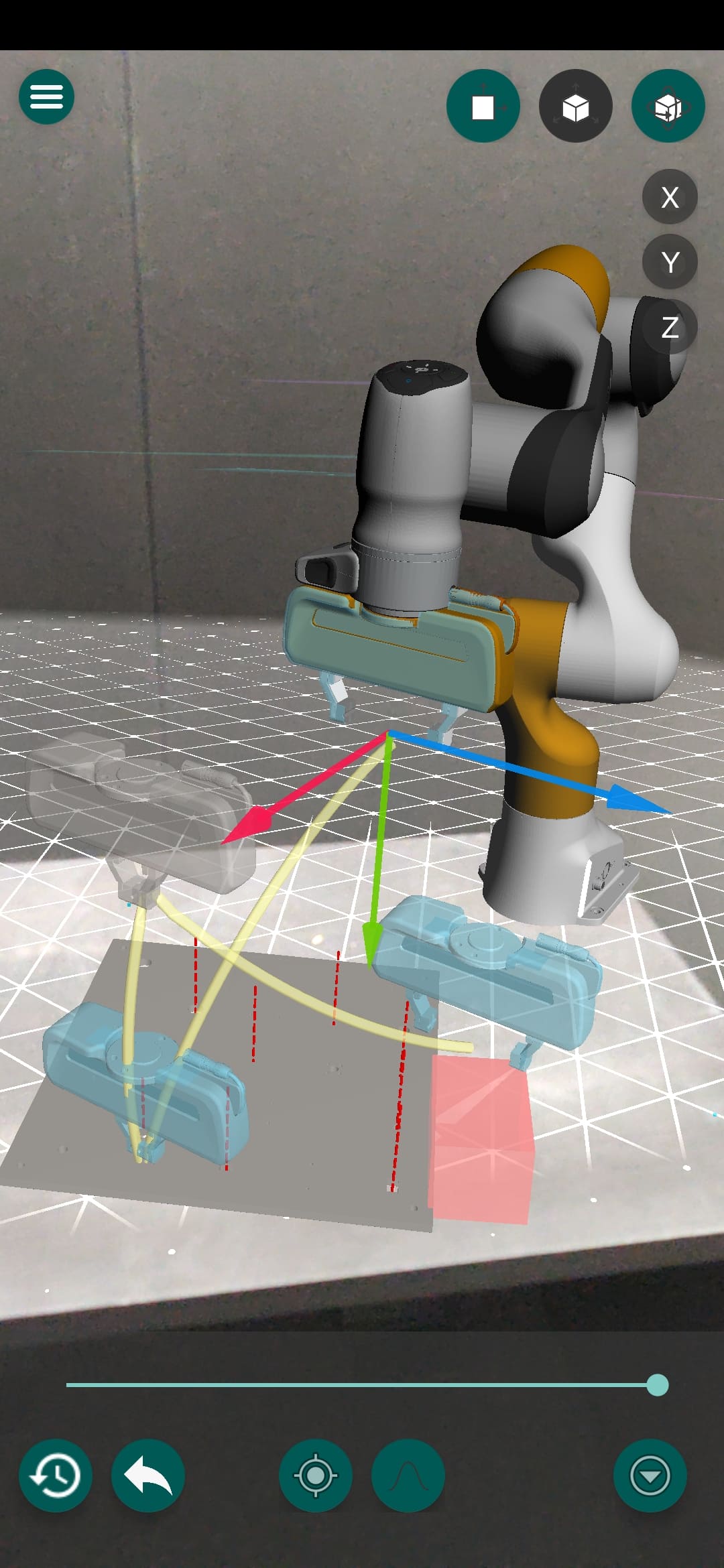}}
    \hfill
    \subfloat[\label{fig_gaussian}]{\includegraphics[width=1.5in]{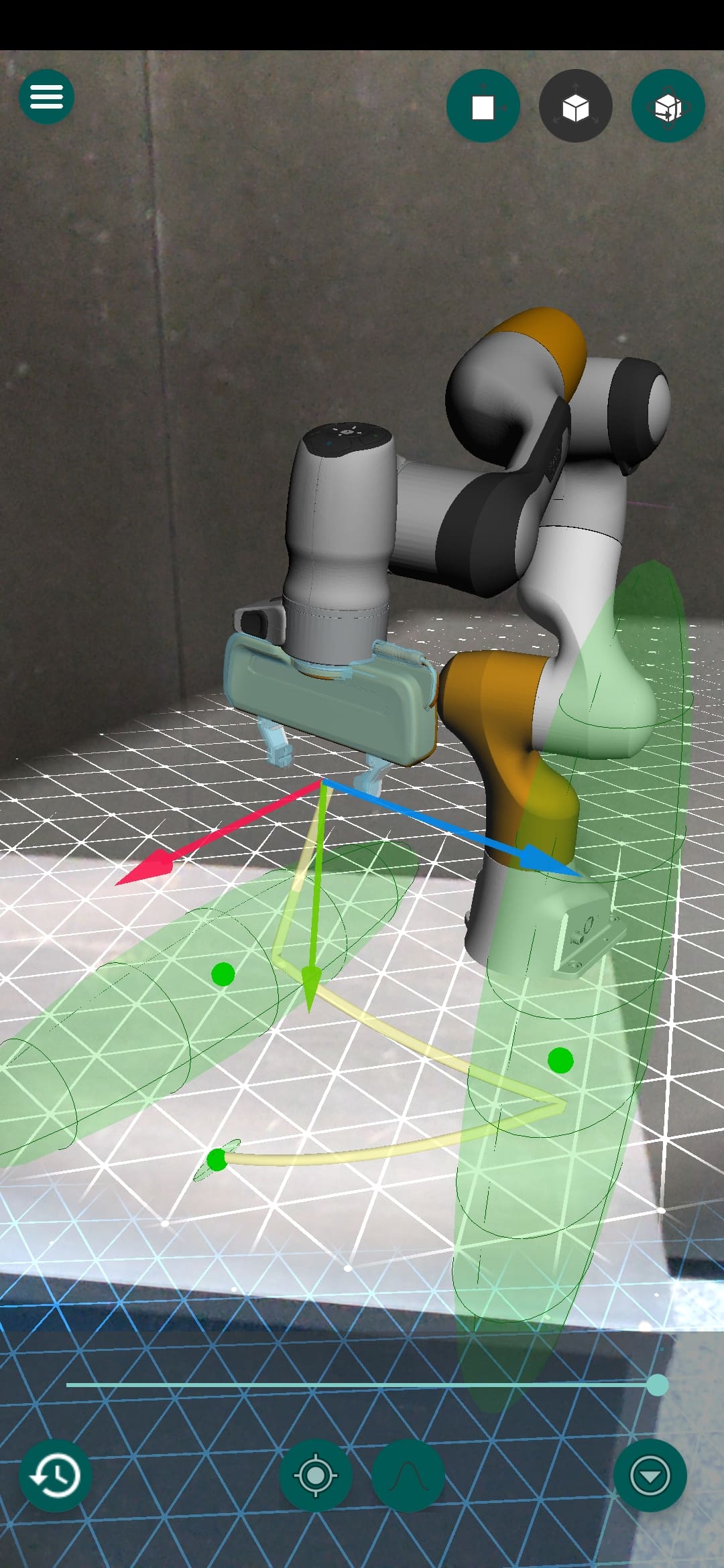}}
    \caption{Trajectory planning with (a) keypoints illustrated by transparent end-effectors and (b) Gaussians (green ellipsoids) where their scale represents the (co)variations allowed along each axis, which are then converted to full precision matrices within the iLQR optimal control technique employed for planning.}
    \label{fig_trajectoryPlanning}
\end{figure}

The third option enables the user to move the end-effector constrained on a 2D plane defined by the device's orientation, as seen in Fig. \ref{fig_planeMotion}. This option provides a more intuitive way to interact with the robot from the perspective of the user who may also be moving. The final option is the direct control of the joints that are highlighted in the interface such as the elbow joint, as illustrated in Fig. \ref{fig_elbowManip}. 
The application also provides the user with the tools to create, save, edit, load and replay end-effector trajectories on the device. To define a trajectory, the user can place keypoints in the workspace by controlling the robot via the techniques described above.
Fig. \ref{fig_exp2} shows four of these workspace keypoints, depicted by partially transparent robot grippers. 

These locations are then sequentially connected by an iterative Linear Quadratic Regulator (iLQR) \cite{IEEEfull:Li_2004} used as a trajectory planner, which determines the corresponding joint trajectory from the task space locations. From this trajectory, an equivalent task space trajectory is computed and depicted with a yellow curve as in Fig. \ref{fig_exp2}. This trajectory and its illustration in the interface are then updated upon the addition, removal or edition of a keypoint.

Additionally, the interface provides the operator with the ability to specify the precision required around each keypoint by replacing them with multivariate Gaussian distributions represented by ellipsoids in the scene (Fig. \ref{fig_gaussian}). The mean of each Gaussian distribution represents the location of the corresponding keypoint, whereas the covariance represents the allowed variance along each principal axis. This view of Gaussian distributions can be exploited within optimal control strategies such as iLQR to define a quadratic cost with the precision matrix being the inverse of the covariance matrix \cite{IEEEfull:Lembono_2021}. This enables the user to directly influence the precision matrices of the trajectory planner by rotating and modifying the scale of each ellipsoid (described in the following section).

As all of these techniques can be performed on the virtual robot or by directly controlling the real robot, the proposed interface is compatible for \emph{online} and \emph{offline} programming.

\subsection{Simulation of virtual objects and workspaces}

In \emph{offline} programming, the user employs a simulator to visualise and control a virtual robot, defining its motions to achieve a task.
Industrial robotics tasks are often defined by the workspace where the robot is located and by the objects and tools that it can interact with.
To this end, we propose to simulate, via the AR interface, workspaces with all their objects/tools available, without requiring the control or presence of the real robot.

\begin{figure}[!t]
\centering
    \subfloat[\label{fig_obsDefinition}]{\includegraphics[width=1.5in]{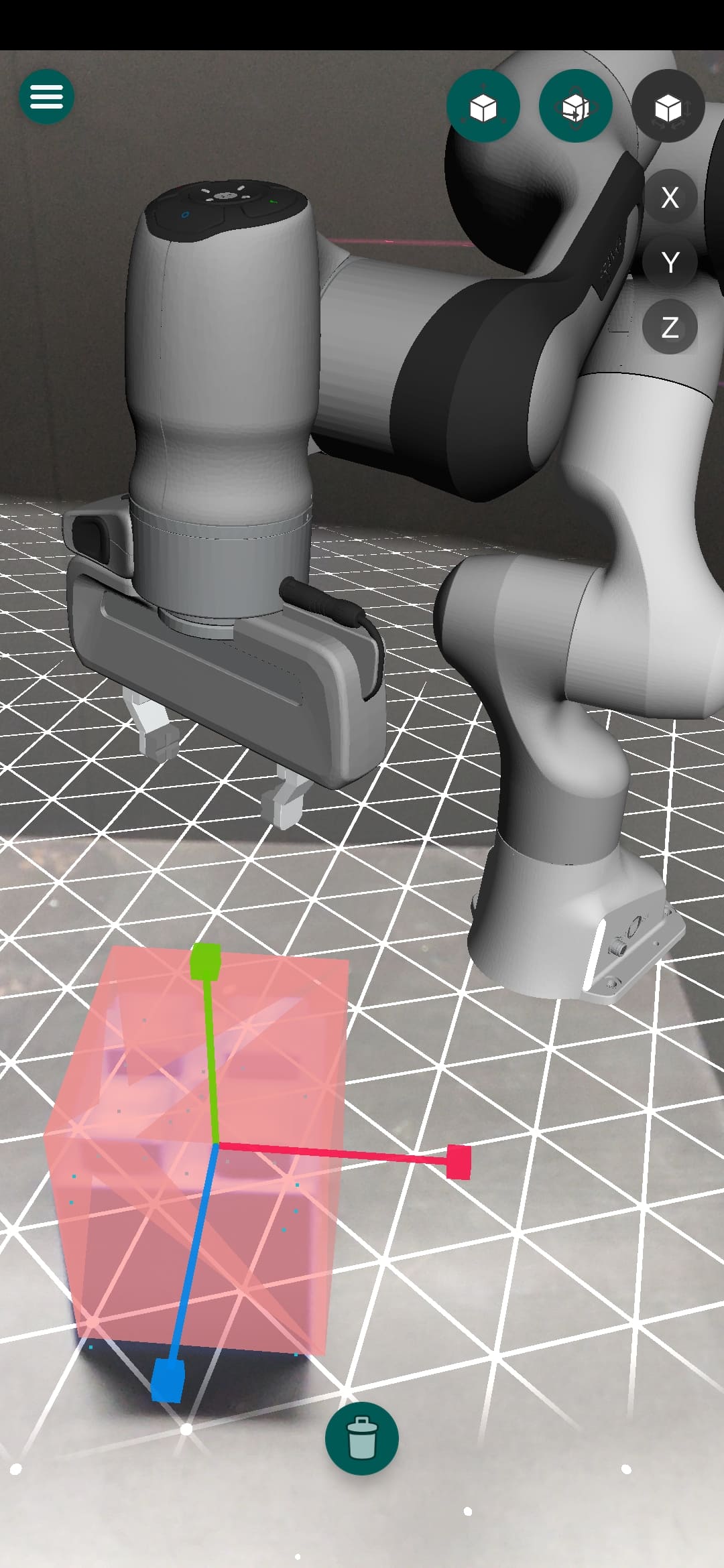}}
    \hfill
    \subfloat[\label{fig_workspace}]{\includegraphics[width=1.5in]{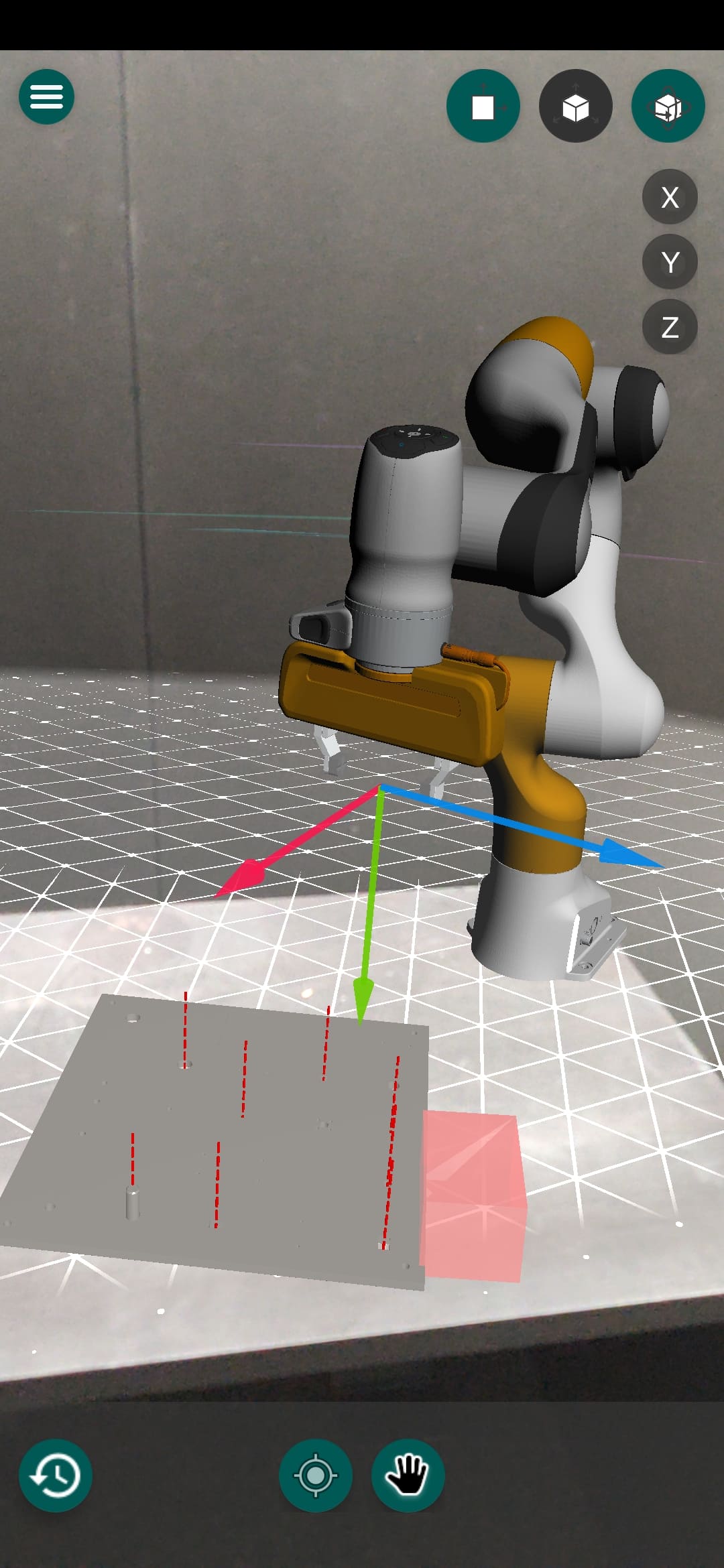}}
    \caption{(a) definition of an obstacle by scaling and positioning a red virtual box to encompass an object; (b) virtual representation of the workspace used for the \emph{offline} condition.}
    \label{fig_virtualDefinitions}
\end{figure}

This provides operators with the ability to design trajectories in a virtual workspace without the need to occupy the real workspace (see Fig. \ref{fig_workspace}).
During the testing phase, adjustments, if required, can be performed directly in the real robot's workspace by first using the \emph{marker calibration} method to superimpose the virtual robot onto the real.
Furthermore, the virtual workspace and the defined trajectories can also be loaded.
Users can therefore edit the trajectory's keypoints (and thus the trajectory itself), by selecting them and manipulating their pose with the same handles described in Section~\ref{section:controlTechniques}.

The interface also allows users to place virtual objects and obstacles (see Fig. \ref{fig_virtComponents}) in the shape of cuboids, spheres and cylinders of modifiable sizes.
The difference between the obstacles (in red in Fig. \ref{fig_virtComponents}) and the objects (in black in Fig. \ref{fig_virtComponents}) is that the former are defined for the user to plan a collision-free trajectory of the robot.
This means that whereas the robot can apply actions on the objects using the physics engine of the interface, we preferred to exclude these interactions on the obstacles to facilitate the programming.
Note that here the collision-free trajectory is defined by the user interaction on the interface and the implementation of an obstacle avoidance planner is left as future work.

These objects and obstacles can be moved using the same methods, described in the previous section, to control the robot's end-effector.
Their scale along each axis can be modified using a similar representation to the one used to translate the end-effector. The creation of such objects and obstacles helps the user to create a virtual workspace that matches at best to the real one (see Fig. \ref{fig_obsDefinition}) and use it as a template such as in Fig. \ref{fig_workspace}. Trajectories can then also be planned in this virtual workspace (Fig. \ref{fig_exp2}) using the functionalities described in the previous section and all the virtual components within the scene (\ie workspace, objects, obstacles and trajectories) can be saved in a file on the device so that they can be reused/adapted for future tasks. 

\section{Experiments}
\label{section:experiments}
 
To evaluate the intuitiveness of the proposed interface and the programming of a robot purely within a smartphone-based AR setup, we conducted a study, with 18 participants, consisting of the programming of a peg disassembly task.
This study was approved by Idiap Research Institute’s Data and Research Ethics Committee.
Our comparison intentionally leaves out traditional ``mouse, keyboard, and screen'' interfaces, as AR approaches have been already compared against such interfaces in~\cite{IEEEfull:Gadre_2019, IEEEfull:Arroyave-Tobon_2015}, showing promising results in terms of speed, accuracy, naturalness of use, and required effort .

\subsubsection{Experimental Setup}
\label{section_expSetup}

The participants programmed a 7-axis Franka Emika robot manipulator, for a peg disassembly task of the National Institute of Standards and Technology (NIST) task board 1 \cite{IEEEfull:Kimble_2020}, see Fig. \ref{fig_expSetup}.
As AR interface, we used an Android smartphone (\textit{6.39-inch} touchscreen display with a resolution of $\textit{1080} \times \textit{2340}$ pixels).
The participants of our study (age range 20--35) had varying knowledge of robotics systems but little to no experience of virtual, augmented and mixed reality.

\subsubsection{Conditions and Protocol}
\label{section:exp_Protocol}

\begin{figure}[!t]
\centering
    \includegraphics[width=3in, height=4in]{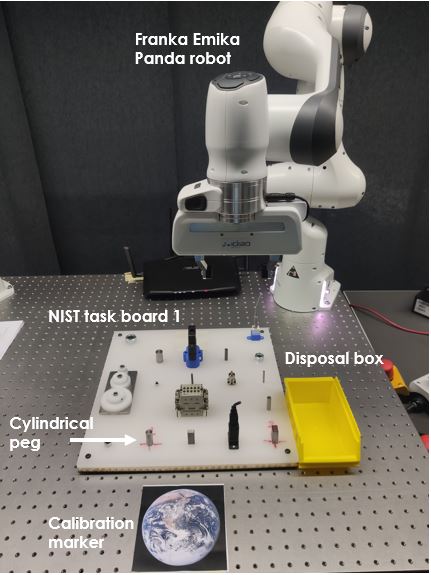}
\caption{Experimental setup with Franka Emika robot, NIST board, calibration marker and disposal box.}
\label{fig_expSetup}
\end{figure}

The participants were asked to perform the disassembly of a cylindrical peg from the NIST task board and place it into a disposal box, as shown in Fig. \ref{fig_expSetup}.
Two conditions were tested in the study: the \textit{online} and the \textit{offline} programming of the disassembly task.

In the \textit{online} condition, the participants directly controlled the real robot's motions from the AR interface and their inputs were immediately translated into robot motions.
Prior to the task, the virtual robot was aligned with the real one based on the marker calibration method described in Section \ref{section:methodology} and the interface was connected to the robot as illustrated in Fig. \ref{fig_markerCalib}.
Furthermore, the real and virtual robots were first set to a default joint configuration after which participants could then start the experiment by displacing the virtual end-effector via the options described in Section \ref{section:controlTechniques}.
The task was considered successfully completed when the peg was correctly disposed inside of the box.
A failure was recorded when the peg was either incorrectly disposed, dropped or an action performed by the user resulted in the robot's safety stop.

In the \textit{offline} condition, participants instead controlled, via the AR interface, a virtual robot acting on a pre-stored virtual copy of the workspace of the \textit{online} condition.
In particular, the participants first placed the virtual workspace on a table different from the one of the real robot (see Fig. \ref{fig_workspace}) and programmed the disassembly of the same cylindrical peg from the board and its placement into a yellow box.
The participants were given 5 minutes of time to define a robot trajectory by means of via-points, as shown in Fig. \ref{fig_exp2}.
The robot trajectory was then sent to the real robot and enacted on the real workspace. If the peg was correctly disposed, the task was considered as successfully completed.
In case of failure, the participants were given one extra minute to perform adjustments to the programmed trajectory in the virtual workspace, according to the visual feedback they observed from the real robot.
This decision was made to provide a fair comparison with respect to the \textit{online} condition where participants experienced real-time feedback while controlling the robot.

We adopted a within-subjects study design, with each participant operating the robot in both the \textit{online} and the \textit{offline} conditions. The order of the conditions was counterbalanced.

Prior to the experiment, the participants were provided with instructions and were given 1-2 minute to familiarise with the AR interface and the disassembly task.

\subsubsection{Questionnaire and Logged Data}
\label{sections:user_questionnaire}

For each condition, we recorded whether the participants were successful in completing the task.
Furthermore, the completion time $\mathbf{t_c}$ of the disassembly task was recorded (\ie seconds between the beginning of the task and its success/failure).
At the end of the experiment, each participant filled a questionnaire with the following 6 Likert scale statements (\textit{1 -- completely disagree, 5 -- completely agree}):
\begin{enumerate}
    \item The interface is easy to understand,
    \item I found the visualisation of the virtual robot useful,
    \item I found it easy to control the robot in real time,
    \item I found it easy to plan trajectories with the virtual robot,
    \item I found the AR interface useful to program the robot,
    \item I would use the AR interface for robot programming.
\end{enumerate}

Each statement included an optional comment section where participants could provide feedback about their experience and provide suggestions on what could be improved.

\begin{figure*}[t]
\centering
\subfloat[\label{fig_exp_res_all}]{\includegraphics[scale=.41]{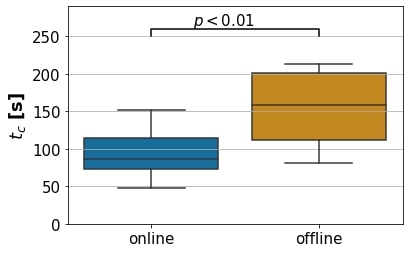}}
\hfill
\subfloat[\label{fig_exp_res_a}]{\includegraphics[scale=.41]{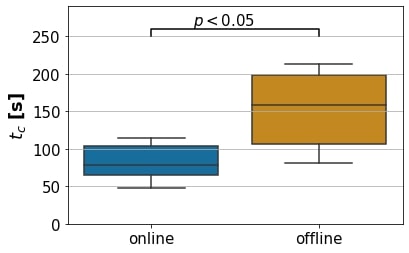}}
\hfill
\subfloat[\label{fig_exp_res_b}]{\includegraphics[scale=.41]{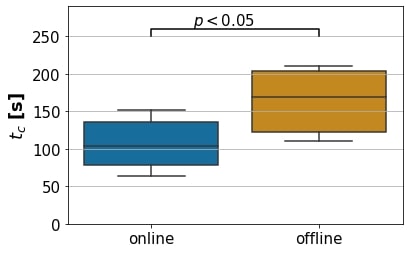}}
\caption{Boxplots of the completion time $\mathbf{t_c}$ acquired during the successful experiments for (a) entire population, (b) \textit{group A} and (c) \textit{group B}. }
\label{fig_exp_time}
\end{figure*}

\section{Results}
\label{section:results}

We hereafter present the analysis of the different metrics collected during the study. When relevant, the data is presented and the analysis is performed by separating the participants into two groups, with the 9 participants who experienced the \textit{online} condition first and then the \textit{offline} condition assigned to \textit{group A}.
The other 9 participants who experienced the conditions in the reverse order were assigned to \textit{group B}.

\begin{table}[b]
\centering
\caption{Success rate on the disassembly task, reported separately by groups and conditions.}
\begin{tabular}{l|ccc} 
\toprule
 & \textbf{\emph{Online} condition} & \textbf{\emph{Offline} condition} & \textbf{Pop. size}  \\ 
\hline
\textit{Group A}         & 100\%            & 78\%           & 9                   \\
\textit{Group B}        & 100\%             & 67\%           & 9                   \\
Overall         & 100\%             & 72\%           & 18                  \\
\bottomrule
\end{tabular}
\label{tbl_exp_acc}
\end{table}

We first computed the success rate of each group for each experiment as summarised in Table \ref{tbl_exp_acc}.
Our hypothesis entering the study was that the success rate in the \textit{offline} condition for the participants in \textit{group A} (that first operated the real-robot in the \textit{online condition}) should be higher with respect to what observed for \textit{group B}.
The hypothesis was motivated by the fact that participants from \textit{group A} have a better understanding of the application and the setup when performing this experiment compared to those of \textit{group B}.
Although we observed a 11\% difference in success rate, no statistically significant difference was found (test on the Agresti-Coull interval, $ p > .05 $).

For the completion time $\mathbf{t_c}$, the descriptive statistics are presented in Fig. \ref{fig_exp_time} as boxplots.
We tested the data for normality with the Shapiro-Wilk test, rejecting the null hypothesis ($ p < .05 $).
For each subset, we therefore ran a non-parametric Wilcoxon signed-rank test for differences between conditions.
Statistically significant differences were found, both for the overall population ($p < .01$), as well as for \textit{groups A} and \textit{B} ($p < .05$).

As expected, we notice a reduction in the time taken to program the task \emph{online} with respect to the \emph{offline} method, as well as a higher success rate.
Based on user feedback, we believe this to be mainly due to the lack of sensory feedback in the virtual environment: the participants had an hard time determining when the centre of the gripper was aligned with the peg that was to be disassembled and when the peg was above the disposal box.

\subsection{User feedback}

\begin{figure}[b]
\centering
    \includegraphics[width=3.5in]{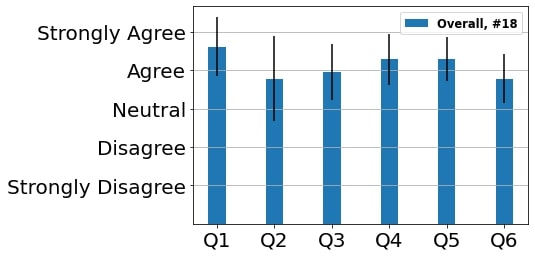}
\caption{Average user ratings for the assertions provided in Section \ref{sections:user_questionnaire}.}
\label{fig_quest_res_all}
\end{figure}

The scores of the Likert scale statements presented in Section \ref{sections:user_questionnaire} are visualised in Fig. \ref{fig_quest_res_all} for the whole study population.
As for the completion time $\mathbf{t_c}$, we looked for differences of questionnaire scores between the two groups with a Mann-Whitney U test, finding however no statistically significant differences ($p > .05$).

Overall, the results indicate that the participants perceived the proposed AR interface as easy to understand and useful for robot programming.
However, out of the 18 participants, only 7 participants stated that they would use the interface for programming industrial robots.
While praising the easiness and quickness of use of the interface, the rest of the participants raised concerns about the interface's lack of accuracy, especially for manufacturing tasks such as insertion.
Most participants mentioned how it was difficult to perform precise robot motions by means of the dragging motion on the smartphone screen.

On the other hand, almost all participants stated that the virtual workspace provides a good representation of the real workspace and that the interface offers an intuitive way of planning trajectories as errors can easily be visualised and corrected.

\section{Discussion}
\label{section:discussion}

The results presented in Section~\ref{section:results} indicate a promising success rate for both \emph{online} and \emph{offline} programming with completion times $\mathbf{t_c}$ ranging from 1 to 4 minutes.
Nevertheless, during the study we observed how certain aspects of the interface could have hindered the participants in accomplishing the task.

A general problem of AR interfaces is their weakness in the estimation and visualisation of depth \cite{IEEEfull:Jamiy_2019}; a limitation that requires the users to adopt coping strategies such as \eg change their location to obtain different views of the scene and, consequently, a better perception of depth.
In our study, we indeed observed how the participants who adopted such coping strategies performed the disassembly tasks faster and more accurately than their counterparts.
This was especially true for the \emph{offline} condition, where the issue of the AR interface with depth was particularly relevant. 

Another issue indicated by the study's participants was in the perceived lack of accuracy during dragging motions across the screen.
This led to inaccurate or unexpected motions on the real robot, especially when the translation axis being manipulated was perpendicular to the surface of the device, leading to substantial changes in robot pose for negligible amounts of input on the screen.
A solution to this issue would be to automatically disable the control of the robot along the axis being employed when the aforementioned condition is met.
While avoiding the problem of unexpected and potentially dangerous robot motions, this solution would also encourage the users to move around the workspace, in order to regain control of a certain axis, indirectly addressing the aforementioned issue of depth perception.

Furthermore, the user feedback also provided insights into various potential areas of research such as using the interface to control the real robot remotely, in a similar fashion to remote teleoperation.
Additionally, some participants stated that they would consider using the AR interface not only as a control tool but also as a visualisation/monitoring tool.
As our interface is ROS-enabled, readings from a multitude of sensors could be visualized in our interface as easily as in other visualization software, like \eg RViz.

Finally, in a Learning from Demonstration scenario \cite{IEEEfull:Calinon_2019,IEEEfull:Argall_2009}, the presented interface could be used to inspect the quality of the demonstrated trajectories or to visualise their variability with aptly placed Gaussians, as shown in~\cite{IEEEfull:Gradmann_2018}.

\section{Conclusion \& Future work}
\label{section:conclusion}

We presented an augmented reality interface for smartphones and/or tablets, enabling users to control a robot in real-time, to program it \emph{offline} as well as to model a workspace by means of virtual objects.
The proposed interface aims to provide operators programming industrial robots with an alternative to common \emph{online} programming methodologies, for which the presence of a real robot is not required and pre-stored workspaces and trajectories can be adapted to re-program a robot. 

Future work will address the challenge of performing more accurate motions. 
We will investigate the option of adapting the sensitivity of the dragging action on the interface, giving the required precision to the user when felt necessary.
Also, we will test the option of switching from the use of dragging motions to a button based system where users can affect the displacement along each axis individually.
This would enable users to perform rapid actions via the current dragging motions on the screen and then switch to the button based system when more accuracy is required.
Such solutions would therefore enable users to tailor the interface to their preferences and provide users who are less experienced with augmented reality with alternative options to control the robot.

Furthermore, as reported by a number of participants, the size and the tactile nature of the device's screen may have a major impact on a user's capacity to program, especially when the task at hand requires the user a careful inspection of the environment.
Future work will explore this design space of our interface and its impact on the interface's usability.

Finally, the ARCore software development kit currently lacks the ability to detect 3D objects.
Having such a feature would allow to automatically place virtual objects or obstacles, resulting in a faster creation of the virtual workspace.
We plan to integrate a 3D object detection pipeline in the interface to provide this capability.
Motivated by the promising results and the participants' feedback presented in Sections \ref{section:results} and \ref{section:discussion}, we also plan to investigate possible extensions of the approach to other robot applications beyond manufacturing.

\end{document}